\documentclass[10pt,twocolumn,letterpaper]{article}
\pdfoutput=1
\usepackage{cvpr}
\usepackage{times}
\usepackage{epsfig}
\usepackage{graphicx}
\usepackage{amsmath}
\usepackage{amssymb}
\usepackage{multirow}
\usepackage{booktabs}
\usepackage{algorithm}
\usepackage[noend]{algpseudocode}
\makeatletter
\def\BState{\State\hskip-\ALG@thistlm}
\makeatother

\usepackage{color}
\usepackage{xspace}

\usepackage[pagebackref=true,breaklinks=true,letterpaper=true,colorlinks,bookmarks=false]{hyperref}

\newcommand{\dttw}{D${}^3$TW\xspace}
\newcommand{\eqnref}[1]{{Eq.\ \eqref{eq:#1}}}

\cvprfinalcopy 

\def\cvprPaperID{1466} 

\ifcvprfinal
\begin{document}

\title{D${}^3$TW: Discriminative Differentiable Dynamic Time Warping \\ for Weakly Supervised Action Alignment and Segmentation}
\vspace{-2mm}
\author{Chien-Yi Chang, De-An Huang, Yanan Sui, Li Fei-Fei, Juan Carlos Niebles\\
Stanford University, Stanford, CA 94305, USA\\
}

\maketitle
\thispagestyle{empty}

\begin{abstract} 
We address weakly supervised action alignment and segmentation in videos, where only the order of occurring actions is available during training.
We propose Discriminative Differentiable Dynamic Time Warping (D${}^3$TW), the first discriminative model using weak ordering supervision. The key technical challenge for discriminative modeling with weak supervision is that the loss function of the ordering supervision is usually formulated using dynamic programming and is thus not differentiable. We address this challenge with a continuous relaxation of the min-operator in dynamic programming and extend the alignment loss to be differentiable. The proposed \dttw innovatively solves sequence alignment with discriminative modeling and end-to-end training, which substantially improves the performance in weakly supervised action alignment and segmentation tasks. We show that our model is able to bypass the degenerated sequence problem usually encountered in previous work and outperform the current state-of-the-art across three evaluation metrics in two challenging datasets.
\end{abstract}
\section{Introduction}
\label{sec:introduction}

Video action understanding has gained increasing interest over recent years because of the large amount of video data.
In contrast to fully annotated approaches~\cite{kuehne2014language,rohrbach2012database,yeung2015every} which require annotations of the exact start and end time of each action, \emph{weakly supervised} approaches~\cite{ding2018weakly,huang2016connectionist,richard2018neuralnetwork,bojanowski2014weakly,kuehne2017weakly} significantly reduce the required annotation effort and improve the applicability to real-world data. In particular, we focus on one type of weak label commonly referred to as \emph{action order} or \emph{transcript}, which uses an ordered list of actions occurring in the video as supervision.

The major challenge of using only the action order as supervision is that the ground truth target, frame-wise action label is not available at training time.  Previous work resorts to using a variety of surrogate loss functions that maximize the posterior probability of the weak labels or the action ordering given the video. However, as shown in~\cite{huang2016connectionist}, using surrogate loss functions can easily lead to degenerated results that align some occurring actions to a single frame in the video. Such degenerated results are far from the ground truth we desire because each action usually spans many frames during its execution. While previous works have attempted to address this challenge using frame-to-frame similarity~\cite{huang2016connectionist}, fine-to-coarse strategy~\cite{richard2017weakly}, and segment length modeling~\cite{richard2018neuralnetwork}, these approaches still consider the degenerated results that align to single frames as valid solutions subject to the surrogate loss functions.

\begin{figure}[tb]
\centering
   \includegraphics[width=1.0\linewidth]{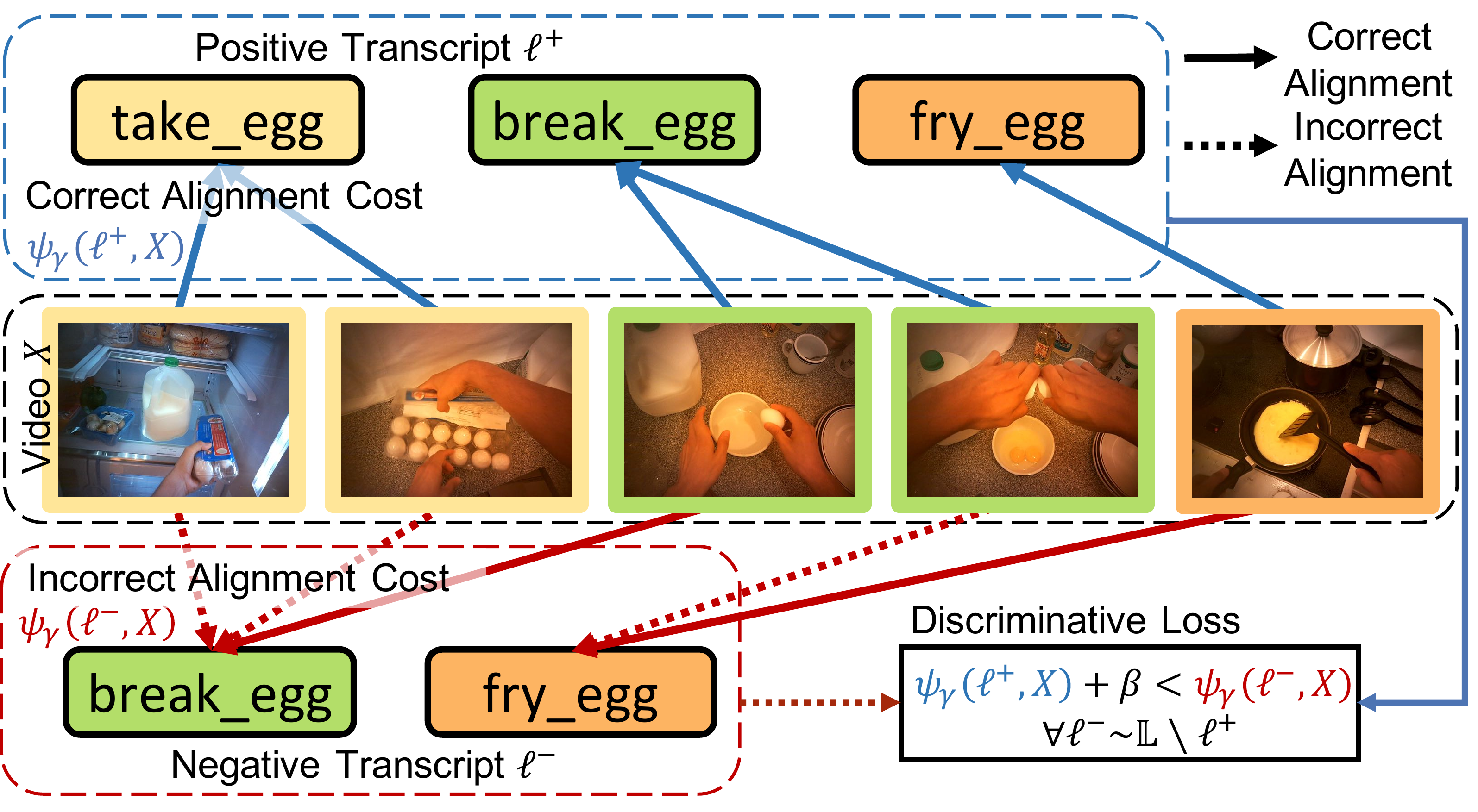}
  \caption{
   We use only the ordered list of actions or the transcript as weak supervision for training. 
   This setting is challenging as the desired output is not available at training. We address this challenge by proposing the first discriminative model for this task. The cost $\psi_\gamma(\ell^+, X)$ of aligning the video $X$ (middle) to the ground truth or positive transcript $\ell^+$ (top) should be smaller than that of the negative transcript $\ell^-$ (bottom) that are randomly sampled.
   }
   \vspace{-1mm}
\label{fig:fig1}
\end{figure}

{The main contribution of this paper is to address the challenge by proposing the first \emph{discriminative} model using order supervision. As illustrated in Figure~\ref{fig:fig1}, the idea is that the probability of having the correct alignment with the positive or ground truth transcript should be higher than that of negative transcripts. In contrast to previous works that only maximize the posterior probability of the weak labels~\cite{huang2016connectionist,richard2017weakly,richard2018neuralnetwork}, our discriminative formulation does not suffer from the degenerated alignment as it is no longer an obvious and trivial solution to the newly proposed discriminative loss. } Further, minimizing the discriminative loss directly contributes to the improvement of our target in contrast to previous work. Similar ideas have been studied in other research areas, such as multiple-instance learning for image tagging, and have been shown to be successful \cite{wu2015deep}.

While the idea of applying discriminative modeling to weakly supervised action labeling problem is seemingly intuitive, the key technical challenge is that the computation of loss functions in previous methods usually involves non-differentiable structural prediction algorithms such as dynamic programming (DP). We address this challenge by proposing Discriminative Differentiable Dynamic Time Warping (\dttw), where we directly optimize for better outputs by minimizing a discriminative loss function obtained by continuous relaxation of the minimum operator in DP \cite{mensch2018differentiable}. The use of \dttw allows us to incorporate the advantage of discriminative modeling with structural prediction model, which was not possible in previous approaches. 

We evaluate \dttw on two weakly supervised tasks in two popular benchmark datasets, the Breakfast Action~\cite{kuehne2014language} and the Hollywood Extended~\cite{bojanowski2014weakly}. The first task is action \emph{segmentation}, which refers to predicting frame-wise action labels, where the test video is given without any further annotation. The second task is action \emph{alignment}, as proposed in \cite{bojanowski2014weakly}, which refers to aligning a test video sequence to a given action order sequence.
We show that our \dttw significantly improves the performance on both tasks.

In summary, our key contributions are: 
(i) We introduce the first discriminative model for ordering supervision to address the degenerate sequence problem. 
(ii) We propose \dttw, a novel framework that incorporates the advantage of discriminative modeling and end-to-end training for structural sequence prediction with weak supervision. 
(iii) We apply our method in two challenging real-world video datasets and show that it achieves state-of-the-art for both weakly supervised action segmentation and alignment.
\section{Related Works}
\label{sec:relatedworks}

\noindent\textbf{Action Recognition and Segmentation.} 
Action recognition has been an important task for video understanding~\cite{heilbron2015activitynet,pirsiavash2014parsing,sener2015unsupervised,vo2014stochastic}. As performances on trimmed video datasets advance~\cite{heilbron2015activitynet,carreira2017quo}, recent focus of video understanding has shifted towards longer and untrimmed video data, such as VLOG~\cite{fouhey2018lifestyle}, Charades~\cite{sigurdsson2016hollywood}, and EPIC-Kitchens~\cite{Damen2018EPICKITCHENS}. This has led to the development of action segmentation approaches~\cite{lea2016temporal,sigurdsson2017asynchronous,yeung2015every} that aim to label every frame in the video and not just to classify trimmed video clips. Our goal is also to densely label each frame of the video, but without the dense supervision for training. 

\vspace{1mm}
\noindent\textbf{Weakly Supervised Learning in Vision.}  For images, weakly supervised learning has been studied in classification~\cite{wu2015deep,mahajan2018exploring}, semantic segmentation~\cite{zhang2015weakly}, object detection~\cite{kumar2016track}, and visual grounding~\cite{karpathy2015deep,xiao2017weakly}. The ordering constraint has been used widely as weak supervision in videos~\cite{bojanowski2014weakly,bojanowski2015weakly,ding2018weakly,huang2016connectionist,richard2017weakly,richard2018neuralnetwork}. The closest to our work is the NN-Viterbi~\cite{richard2018neuralnetwork}, where the it combines a neural network and a non-differentiable Viterbi process to learn from ordering supervision iteratively. In contrast, the proposed \dttw is end-to-end differentiable and uses discriminative modeling to directly optimize for the best alignment under ordering supervision.

\vspace{1mm}
\noindent\textbf{Using Language as Supervision for Videos.} As the ordering supervision can be automatically extracted from language, our work is related to using language as supervision for videos. The supervision usually comes from movie scripts~\cite{duchenne2009automatic,bojanowski2015weakly,zhu2015aligning} or transcription of instructional videos~\cite{alayrac2015learning,sener2015unsupervised,malmaud2015s,huang-buch-2018-finding-it}. Unlike these approaches, we assume the discrete action labels are already extracted and focus on leveraging the ordering information as supervision.

\vspace{1mm}
\noindent\textbf{Continuous Relaxation.} 
Our \dttw  is related to recent progress on continuous relaxation of discrete operations, including theorem proving~\cite{rocktaschel2017end}, softmax function~\cite{jang2017categorical}, logic programming~\cite{evans2018learning}, and dynamic programming~\cite{mensch2018differentiable, cuturi2017soft}. We use the same principle and further enable discriminative modeling of dynamic programming based alignment.

\section{Method}
\label{sec:method}

\begin{figure}[tb]
\centering
   \includegraphics[width=0.95\linewidth]{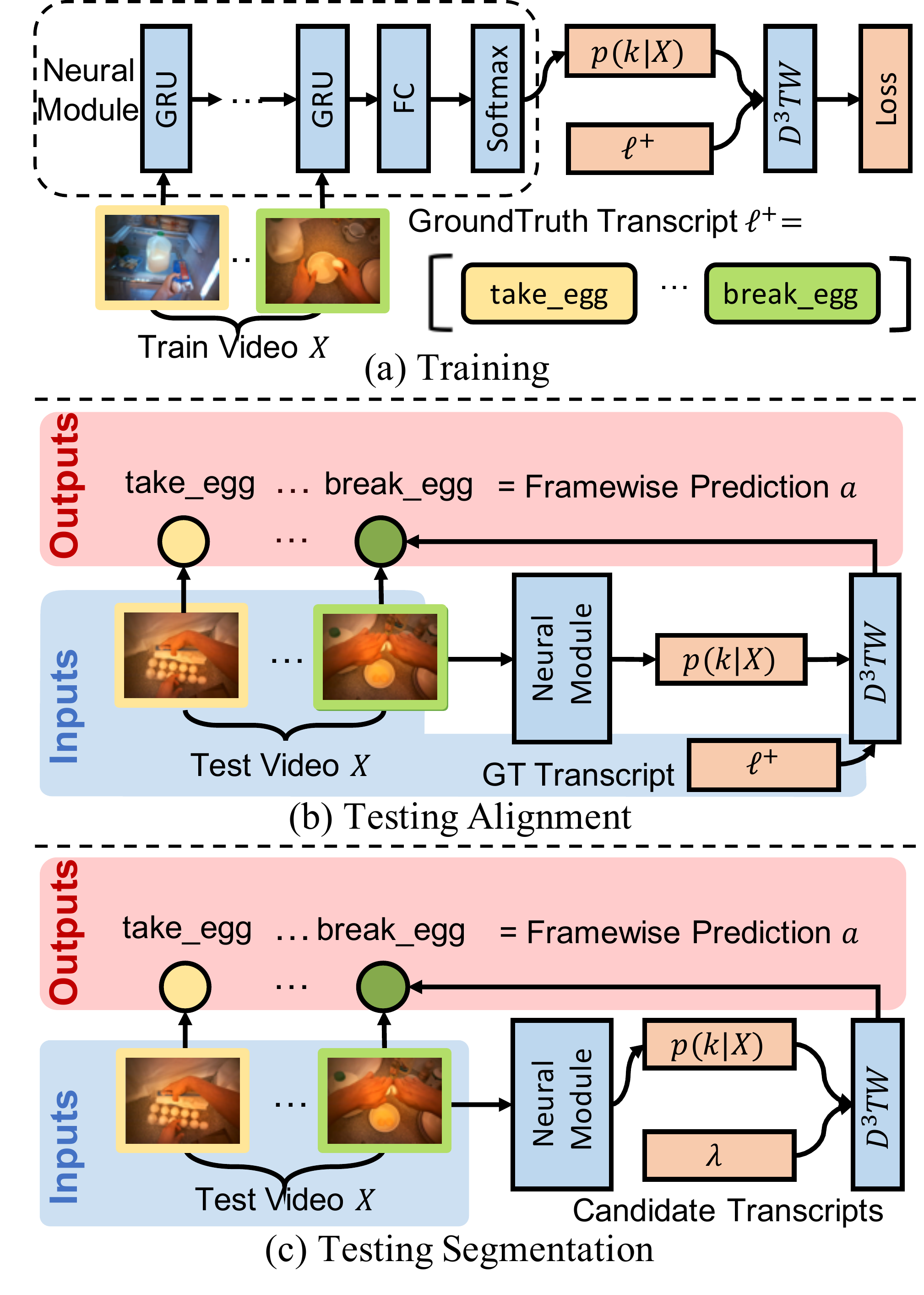}
   \vspace{-2mm}
   \caption{
      \textbf{(a)} During training, only the transcript $\ell^+$ is given. The input video is first forwarded through a GRU to generate the posterior probabilities $p(k|X)$ of each action for each frame. \dttw is a discriminative model with a fully differentiable loss function, which allows us to learn $p(k|X)$ via backpropagation and sets our approach apart from previous work. \textbf{(b)} For alignment, at test time our \dttw loss can directly be used to align the given transcript $\ell^+$ with the video sequence.  \textbf{(c)} For segmentation, at test time no transcript is given. We reduce segmentation to alignment by aligning the video to a set of candidate transcripts $\lambda$ and output the best candidate as the segmentation result. 
   }
   \vspace{-1mm}
\label{fig:system_v2}
\end{figure}

Our goal is to learn to temporally align and segment video frames using only weak supervision, where only the order of occurring actions is available at training.
The major challenge for weakly supervised problem is that the ground truth target, i.e., frame-wise action labels are not available at training. We address this challenge by proposing Discriminative  Differentiable  Dynamic Time Warping (\dttw), which is to our best knowledge, the first discriminative modeling framework with ordering supervision. The use of discriminative modeling and differentiable dynamic programming sets our approach apart from previous work that involves non-differentiable forward-backward algorithms ~\cite{graves2006connectionist, huang2016connectionist, richard2018neuralnetwork} and dramatically alleviates the problem of degenerated alignments that aligns each action label to a single frame. Figure~\ref{fig:system_v2} shows the outline of our model.

In the following, we describe our framework in detail, starting with the problem statement. We then define our model and show how it can be used at test time.

\subsection{Weakly Supervised Action Learning}

We start with the definition of the weakly supervised action alignment and segmentation. Here the weak supervision means that only the \emph{transcript}, or an ordered list of the actions is provided at training time. A video of frying eggs, for example, might consist of taking eggs, breaking eggs, and frying eggs. While the full supervision would provide the fine-grained temporal boundary of each action, in our weakly supervised setup, only the action order sequence \verb|[take_egg, break_egg, fry_egg]| is given.

We address two tasks in this paper: action \emph{segmentation} and action \emph{alignment}. We aim to learn both with weak supervision. As shown in Figure~\ref{fig:system_v2}(b) and (c), the difference between the two tasks is that at test time, action alignment uses both transcript and test video frames as input, while action segmentation only requires  test video frames as inputs. We observe that action segmentation can be formulated as an action alignment task given a set of possible transcripts at test time. We will first explain how to tackle action alignment using weak supervision, and explain how action segmentation can be reduced to the action alignment problem.

Formally, given an input sequence of video frames $X = [x_1, \cdots, x_{T}] \in \mathbb{R}^{d \times T}$, the goal of action alignment is to predict an output alignment sequence of frame-wise action labels $\hat a = [\hat a_1, \cdots, \hat a_{T}] \in \mathcal{A}^{1\times T}$, under the constraint that $a_i$ follows the action order in the transcript  $\ell^+ = [\ell^+_1, \cdots, \ell^+_{L}]\in \mathcal{A}^{1\times L}$. Here, $\mathcal{A}$ is the set of possible actions. In other words, we want to learn a model $f(X, \ell^+) = \hat{a}$. The key challenge of weak supervision is that we only have the inputs $(X, \ell^+)$ as supervision for training $f(\cdot)$ without access to the ground truth action labels $a^+_{1:T}$.

For action segmentation, we observe that segmentation can be formulated as alignment given a set of possible transcripts. Formally, given a set of possible transcripts $\mathbb{L}$, let $\Psi(a, X) \in \mathbb{R}$ be a score function that measures the goodness of predicted action labels $a$ given input video $X$, action segmentation task can be solved by exhaustive search
\begin{equation}
\vspace{-1mm}
\label{eq:unified}
\small
\hat a = \underset{a = f(\ell,X), \ell \sim \mathbb{L}}{\mathrm{arg max}} \Psi \left (a, X\right).
\end{equation}
This finds the candidate transcript $\ell$ that gives the best alignment measured by $\Psi(\cdot, X)$ for transcripts in $\mathbb{L}$.

\subsection{Discriminative Differentiable DTW (\dttw)}

We have discussed what is weakly supervised action alignment and how we can solve action segmentation based on alignment. Now we discuss how we use discriminative modeling to learn a model that aligns the transcript $\ell^+$ and the video frames $X$ using just $\ell^+$ and $X$ at training.

We pose action alignment as a Dynamic Time Warping  (DTW)~\cite{sakoe1978dynamic} problem, which has been widely applied to sequence alignment in speech recognition. Given a distance function $d(\ell^+_i, x_j)$ that measures the cost of aligning the frame $x_j$ to a label in the transcript $\ell^+_i$, DTW uses dynamic programming to efficiently find the best alignment that minimizes the overall cost. The key challenge of weakly supervised learning is that there is no frame-to-frame alignment label to train this distance function $d(\ell^+_i, x_j)$. We address this challenge by proposing Discriminative Differentiable Dynamic Time Warping (\dttw), which allows us to learn $d(\ell^+_i, x_j)$ using only weak supervision. In the following, we will first discuss how we formulate video alignment as DTW and next how we learn the distance function $d(\ell^+_i, x_j)$ using \dttw. 

\subsubsection{Video Alignment as Dynamic Time Warping}
\label{sec:vid_align}

Given two sequences $\ell$ and $X$ of lengths $L$ and $T$ corresponding to the transcript and the video, we define $\mathcal{Y} \subset \{0,1\}^{L\times T}$ to be the set of possible binary alignment matrices. Here $\forall Y \in \mathcal{Y}$, $Y_{ij} = 1$ if video frame $x_j$ is labeled as $\ell_i$ and $Y_{ij} = 0$ otherwise. We impose rigid constraints on eligible warping paths based on the observation that each video frame can only be aligned to a single action label, such that the alignment from $X$ to $\ell$ is strictly one-to-one. In other words, $\mathcal{Y} \subset \{0,1\}^{L\times T}$ is the set of binary matrices with exactly $T$ nonzero elements and column pivots. Given an alignment matrix $Y$, we can derive its corresponding action label $a_{1:T}$ as: $a_j = \ell_i$, if $Y_{ij} = 1$.

Given the constraints on the eligible alignments, the goal of DTW is to find the best alignment $Y^* \in \mathcal{Y}$ 
\begin{equation}
\vspace{-1mm}
\label{eq:allpath}
\small
Y^{*} = \underset{Y \in \mathcal{Y}}{\mathrm{argmin}} \langle Y, \Delta(\ell, X) \rangle,
\end{equation}
that minimizes the inner product between the alignment matrix $Y$ and the distance matrix $\Delta(\ell, X)$ between transcript $\ell$ and video $X$, where $\Delta (\ell, X) := [d(\ell_i, x_j)]_{ij} \in \mathbb{R}^{L\times T}$.

\begin{figure}[tb]
\centering
   \includegraphics[width=1.0\linewidth]{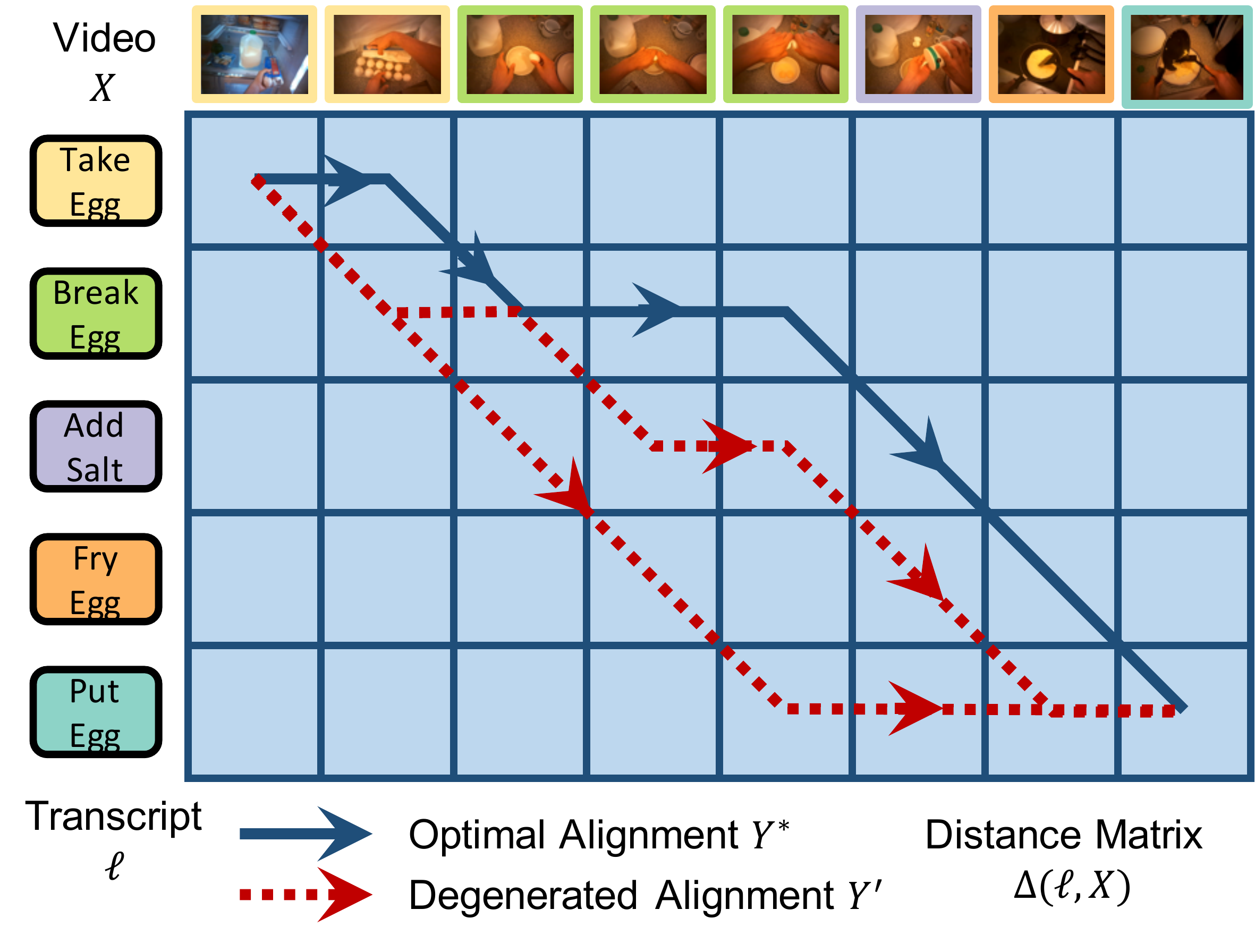}
  \caption{
  Dynamic Time Warping formulation for video alignment.  The $5\times8$ colored grid represents distance matrix $\Delta(\ell, X)$. Here we use a trellis diagram to show the computational graph of the optimal transcript-video alignment $Y^*$ as defined in Eq. \eqref{eq:allpath}. Bellman recursion guarantees that $\langle Y^*, \Delta \rangle \le \langle Y', \Delta \rangle, \forall Y' \in \mathcal{Y}$ and the action order in the transcript is strictly preserved.
  }
   \vspace{-1mm}
\label{fig:fig3}
\end{figure}

Given the distance function $d(\ell_i, x_j)$, we can solve ~\eqnref{allpath} using dynamic programming. A simplified example of such process is illustrated in Figure~\ref{fig:fig3}. Of all paths that connect the upper left entry $\Delta_{11}$ to the lower right entry $\Delta_{LT}$ using only $\longrightarrow$, $\searrow$ moves, $Y^*$ is the optimal alignment that minimizes the alignment cost between transcript sequence and video frames. In this case, we can efficiently obtain the best alignment between video $X$ and transcript $\ell$.

\subsubsection{Discriminative Modeling with Weak Supervision}
We have discussed how we obtain the best alignment $Y^*$ given the distance function $d(\ell_i, x_j)$ using DTW. However, 
the problem remains that how can we learn this distance function without access to the ground truth alignment.

An approach used in prior work~\cite{huang2016connectionist,richard2017weakly,richard2018neuralnetwork} maximizes the probability of the video $X$ given the transcript $\ell$:
\begin{equation}
\vspace{-1mm}
\label{eq:likelihood}
\small
    p(X|\ell) = \sum_a \prod_t p(x_t|a_t) p(a_t|\ell),
\end{equation}
where $a_t \in \mathcal{A}$ is the action label for frame $t$. By optimizing the objective in \eqnref{likelihood}, we can learn $p(x_t|k)$, the probability of observing $x_t$ given action $k \in \mathcal{A}$. In order to maximize the probability, we define the distance $d(\ell_i, x_j) = -\log p(x_j|\ell_i)$ as the negative log-likelihood.

One should notice that the alignment $a_t$ in  \eqnref{likelihood} is latent and the number of possible alignments grows exponentially with the length of the video. Therefore, previous work either uses dynamic programming~\cite{huang2016connectionist}, or uses a hard EM approach~\cite{richard2017weakly,richard2018neuralnetwork} to infer $a_t$ and iteratively maximize the objective in \eqnref{likelihood}. The key drawback of such approaches is that they can easily lead to a degenerate or trivial solution as the space of alignments is too large. While one can impose constraints by enforcing heuristic priors on the possible alignments $p(a_t|\ell)$, this does not directly address the drawback that maximizing this objective does not necessarily lead to the correct alignment.

\begin{figure}[tb]
\centering
   \includegraphics[width=0.8\linewidth]{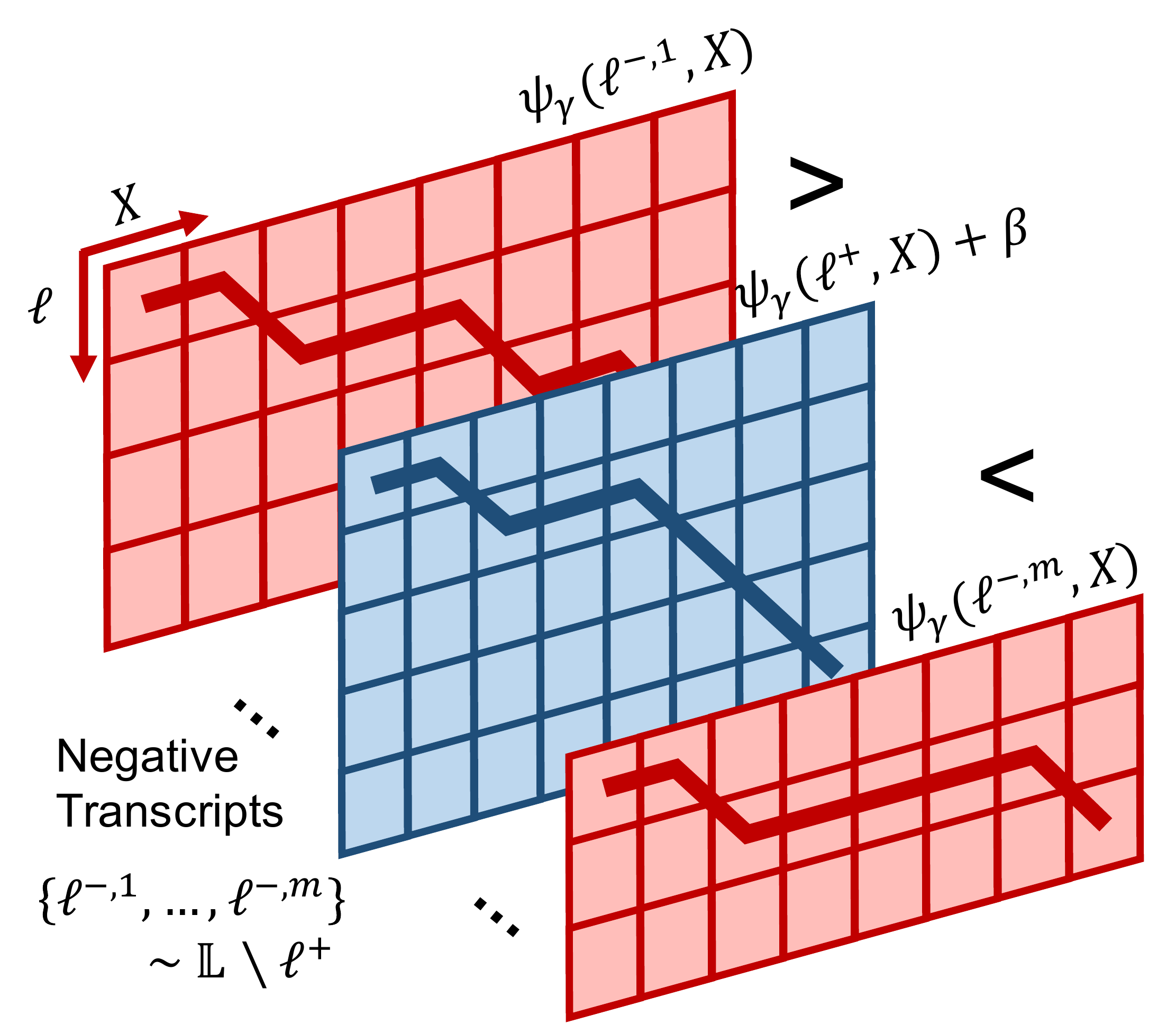}
   \caption{
    We introduce discriminative modeling to weakly supervised action alignment. The loss $\psi_\gamma(\ell^+, X)$ of aligning the video $X$ to the correct transcript $\ell^+$ should be lower than that of any other randomly sample negative transcript $\ell^-$, which prevents degenerated alignments issue commonly seen in previous work.
   }
   \vspace{-1mm}
\label{fig:d3tw}
\end{figure}

Our key insight here is to introduce discriminative modeling to the weak ordering supervision problem. We enforce a discriminative constraint that should hold for any input tuple $(\ell^+, X)$, that 
\begin{equation}
\vspace{-1mm}
\small
    \label{eq:discriminant}
    p(X|\ell^+) > p(X|\ell^-), \forall \ell^- \in \mathbb{L} \setminus \ell^+,
\end{equation}
where the probability of observing the video based on the ground truth or positive transcript $\ell^+$  should always be higher than the probability observing the video from the negative transcript $\ell^-$, as illustrated in Figure~\ref{fig:d3tw}. This discriminative constraint was not explicitly used in previous work. Using the hinge loss with margin $\beta \ge 0$, the loss function can be written as:
\begin{equation}
\vspace{-1mm}
\small
    \label{eq:hinge_prob}
    \sum_{\ell^- \sim \mathbb{L} \setminus \ell^+} \max (p(X|\ell^+) - p(X| \ell^-), \beta).
\end{equation}

\subsubsection{Differentiable Loss with Continuous Relaxation}
\label{sec:method_d3tw}
While the above discriminative modeling is intuitive, the technical challenge is that $p(X|\ell^+)$ and $p(X|\ell^-)$ in \eqnref{hinge_prob} are generally not differentiable with respect to the distance function $d(\ell_i, x_j)=-\log p(x_j|\ell_i)$ we aim to learn. One way of optimizing it is to use hard EM~\cite{richard2017weakly,richard2018neuralnetwork} and iteratively optimize this loss given the current distance function $d(\ell_i, x_j)$. However, hard EM is numerically unstable because it uses a hard maximum operator in its interactions to update model parameters~\cite{mensch2018differentiable}. The key technical contribution of our approach is proposing a continuous relaxation of the DTW-based video alignment loss function. 

Instead of iteratively updating the model parameters by solving \eqnref{allpath} to find the best alignment given the current $d(\ell_i, x_j)$ with hard EM, we can solve the following continuous relaxation:
\begin{equation}
\vspace{-1mm}
\small
\label{eq:dtw}
\psi_{\gamma}(\ell, X) = \min{}_{\gamma} \{\langle Y ,  \Delta (\ell, X)\rangle,  Y \in \mathcal{Y}\}.
\end{equation}
Here $\min_{\gamma}\{\}$ is the continuous relaxation of regular minimum operator regularized by negative entropy $H(q) = -\sum q \log(q)$ with a smoothing parameter $\gamma \ge 0$, such that
\begin{equation}
\vspace{-1mm}
\label{eq:softmin}
\small
\min{}_\gamma \{a_1, \cdots, a_n\} = 
    \begin{cases}
      \min_{i\le n} a_i, & \gamma=0 \\
      -\gamma \log \sum_{i=1}^n e^{-a_i/\gamma}, & \gamma > 0
    \end{cases}.
\end{equation}
This transforms the dynamic programming based DTW loss function into a differentiable one with respect to $d(\ell_i, x_j)$ when $\gamma > 0$. The smoothing parameter $\gamma$ empirically helps the optimization although it does not explicitly convexify the objective function.
The gradient of Eq.~\eqref{eq:dtw} can be derived using the chain rule:
\begin{equation}
\vspace{-1mm}
\label{eq:graddtw}
\small
\nabla_{X} \psi_{\gamma}(\ell, X) = 
\left(\frac{\partial \Delta(\ell, X)}{\partial X}\right)^T  \frac{\sum_{Y\in\mathcal{Y}} e^{-\langle Y, \Delta(\ell, X) \rangle / \gamma}Y}{\sum_{Y\in\mathcal{Y}} e^{-\langle Y, \Delta(\ell, X) \rangle / \gamma}},
\end{equation}
where the second term on the right can be interpreted as the average alignment matrix under the Gibbs distribution $p_\gamma \propto e^{-\langle Y, \Delta(\ell, X) \rangle / \gamma}, \forall Y \in \mathcal{Y}$. Algorithm \ref{algo:dtw} summarizes the procedure for computing $\psi_{\gamma}(\ell, X)$ and its gradient. 

We can interpret $\psi_{\gamma}(\ell, X)$ as the expectation cost over all possible alignments between transcript $\ell$ and video $X$. Its gradient $\nabla_X \psi_\gamma$ can be seen as a relaxed version of the hard alignment $Y^*$ in ~\eqnref{allpath}. With the continuous relaxation in \eqnref{dtw}, we can directly compute the gradient and optimize for \eqnref{hinge_prob}. This addresses the challenge of getting degenerated alignments due to numerically unstable operations in hard EM. By substituting $p(X|\ell)$ in \eqnref{hinge_prob} with our relaxed alignment cost $\psi_\gamma(\ell, X)$, we obtain the discriminative and differentiable loss function $\mathcal{L}_{\mathrm{D}^3\mathrm{TW}}$:

\begin{equation}
\vspace{-1mm}
\label{eq:d3tw}
\small
\mathcal{L}_{\mathrm{D}^3\mathrm{TW}}(\ell^+, X) = \sum_{\ell^- \sim \mathbb{L} \setminus \ell^+} \max (\psi_{\gamma}(\ell^+, X) - \psi_{\gamma}(\ell^-, X), \beta).
\end{equation}

Directly minimizing \eqnref{d3tw} enables our model to simultaneously optimize for finding the best alignment and discriminating the most accurate transcript given the observed video sequence. 
The differentiablity of \eqnref{d3tw} allows gradients to backpropogate through the entire model and fine-tune the distance function $d(\ell_i, x_j)$ for the distance matrix $\Delta(\ell, X)$ in the alignment task with end-to-end training.

\begin{algorithm}[t]
\small
\caption{Compute alignment cost $\psi_\gamma(\ell, X)$ and its gradient $\nabla_X \psi_\gamma(\ell, X)$ }\label{algo:dtw}
\begin{algorithmic}[1]
\State \textbf{Inputs:} $\ell, X$, smoothing parameter $\gamma \ge 0$, distance function $d$

\Procedure{Forward pass}{}
\State $v_{[0,0]} \gets 0$
\State $v_{[:,0]}, v_{[0, :]} \gets \inf$
\For{$i=[1,\cdots,L]; j=[1,\cdots,T]$}
    \State $v_{[i,j]} \gets d_{[i,j]} + \mathrm{min}_\gamma (v_{[i, j-1]}, v_{[i-1, j-1]})$
    \State $q_{[i,j,:]} \gets \nabla\mathrm{min}_\gamma (v_{[i, j-1]}, v_{[i-1, j-1]})$
\EndFor
\EndProcedure
\Procedure{Backward pass}{}
\State $q_{[:,T+1,:]}, q_{[L+1,:,:]} \gets 0$
\State $r_{[:,T+1]}, r_{[L+1, :]} \gets 0$
\State $q_{[L+1, T+1, :]}, r_{[L+1,T+1]} \gets 1$
\For{$j=[T,\cdots,1]; i=[L,\cdots,1]$}
    \State $r_{[i,j]} \gets q_{[i,j+1,1]} r_{[i,j+1]} + q_{[i+1, j+1, 2]} r_{[i+1, j+1]}$
\EndFor
\EndProcedure
\State \textbf{Returns:} $\psi_\gamma = v_{[L, T]}, \nabla_X \psi_\gamma = r_{[1:L,1:T]}$
\end{algorithmic}
\end{algorithm}

\subsubsection{Learning and Inference}
\label{sec:method_learning_inference}
\vspace{1mm}
\noindent\textbf{Distance Function Parameterization.} 
In this paper, we use a Recurrent Neural Network (RNN) with a softmax output layer to parameterize our distance function $d(\ell_i, x_j)$ given video frames as input. Let $Z = [z_1, \cdots, z_T] \in \mathbb{R}^{A \times T}$ be the RNN output at each frame, where $A = |\mathcal{A}|$ is the number of possible actions. $p(k|x_t) = z_t^k$ can be interpreted as the posterior probability of action $k$ at time $t$. We follow~\cite{richard2018neuralnetwork} and approximate emission probability $p(x_t|k) \propto \frac{p(k|x_t) }{p(k)}$, where $p(k)$ is the action class prior. Action class priors are uniformly initialized to $\frac{1}{A}$ and updated after every batch of iterations by counting and normalizing the number of occurrences of each action class that have been processed so far during the training process.

\vspace{1mm}
\noindent\textbf{Inference for Action Segmentation.} 
At test time we want our model to predict the best action labels $a = [a_1, \cdots, a_T]$ given only an unseen test video $X_\mathrm{test} = [x_1, \cdots, x_T]$. We disentangle the action segmentation task into two components: First, we generate a set of candidate transcripts $\lambda = \{\ell^1, \cdots, \ell^m\} \subset \mathbb{L}$ following~\cite{richard2018neuralnetwork}, where $\mathbb{L}$ represents the set of all possible transcripts. 
Then we align each of the candidate transcripts to the unseen test video $X_\mathrm{test}$ to find the transcript $\hat \ell$ that minimizes the alignment cost $\psi_\gamma$:
\begin{equation}
\vspace{-1mm}
\label{eq:testtime}
\small
\hat \ell = \underset{\ell \in \lambda}{\mathrm{argmin}} \; \psi_\gamma (\ell, X_{\mathrm{test}}).
\end{equation}
The predicted alignment $\hat Y$ and associated frame-level action labels $\hat a$ is given by $\nabla \psi_\gamma (\hat \ell, X)$.

\section{Experiments}
\label{sec:experiments}

\begin{figure*}[tb]
\centering
   \includegraphics[width=1.0\linewidth]{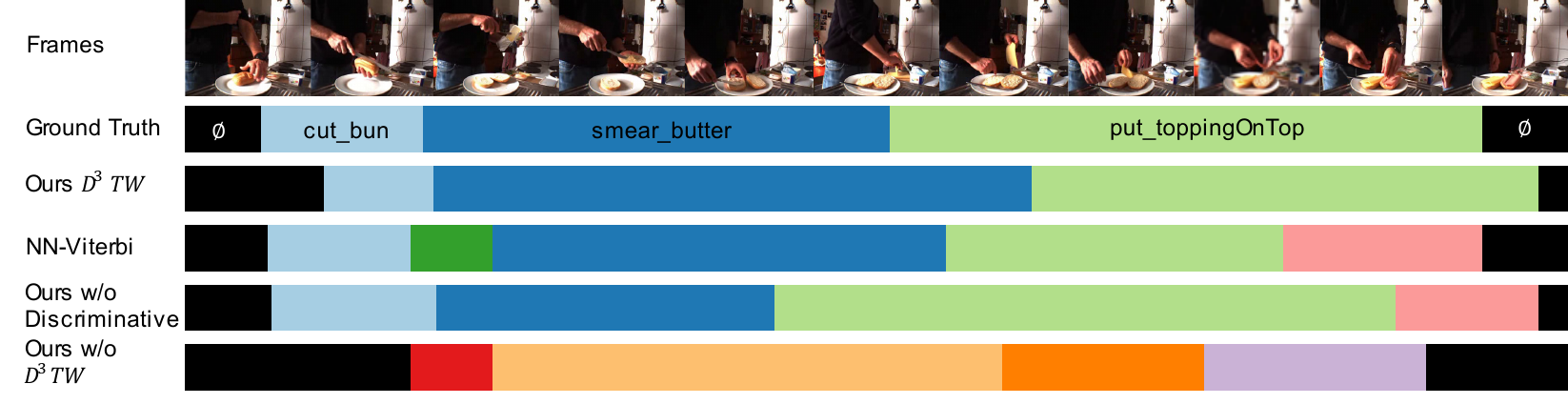}
   \vspace{-4mm}
   \caption{
   Qualitative results on the Breakfast dataset.  Colors indicate  actions and the horizontal axis is time. 
   While both \emph{Ours w/o Discriminative} and \emph{NN-Viterbi} introduce additional actions not appearing in the ground truth, \emph{Ours w/o Discriminative} has better action boundaries because of the differentiable loss.
   \emph{Ours \dttw} is the only model that correctly captures all the occurring actions with discriminative modeling. In addition, this also leads to more accurate boundaries of actions.
   }
   \vspace{-1mm}
\label{fig:vis}
\end{figure*}

The key contribution of D${}^3$TW is to apply discriminative, differentiable, and dynamic alignment between weak labels and video frames. 
In this section, we evaluate the proposed model on two challenging weakly supervised tasks, action \textit{segmentation} and \textit{alignment} in two real-world datasets. In addition, we study how our model's \textit{segmentation} performance varies with more supervision. Through ablation study, we further investigate the effectiveness of the proposed \dttw and compare our approach to current state-of-the-art methods.

\vspace{1mm}
{\noindent \bf Datasets and Features.} {\bf Breakfast Action}~\cite{kuehne2014language} consists of 1,712 untrimmed videos of 52 participants cooking 10 dishes, such as fried eggs, in 18 different kitchens.  Overall, there are around 3.6M frames labeled with 48 possible actions. The dataset has been used widely for weakly supervised action labeling~\cite{ding2018weakly,huang2016connectionist,richard2017weakly,richard2018neuralnetwork}. For a fair comparison, we use the pre-computed features and data split provided by \cite{kuehne2014language}. {\bf Hollywood Extended}~\cite{bojanowski2014weakly} consists of 937 videos containing 2 to 11 actions in each video. Overall, there are about 0.8M frames labeled with 16 possible actions, such as \verb|open_door|. We use the feature and follow the data split in \cite{bojanowski2014weakly} for a fair comparison. 

\vspace{1mm}
{\noindent \bf Network Architecture.} We use single layer GRU ~\cite{graves2005framewise} with 512 hidden units. We optimize with Adam~\cite{kingma2014adam} and cross-validate the hyperparameters such as learning rate and batch size. 

\vspace{1mm}
{\noindent \bf Frame Sub-sampling.} For faster training and inference, we temporally sub-sample feature vectors in Breakfast Action. Following~\cite{huang2016connectionist}, we cluster visually similar and temporally adjacent frames using $k$-means, where $\frac{T}{M}$ centers are temporally uniformly distributed as initialization. We empirically pick $M=20$, which is much shorter than the average length of action ($\sim$400 frames in the Breakfast dataset). No further pre-processing is required for  Hollywood Extended dataset as the feature vectors are already sub-sampled.

\begin{table}
\small

\begin{center}
\begin{tabular}{lcccc}
\toprule
& \multicolumn{2}{c}{Breakfast} & \multicolumn{2}{c}{Hollywood} \\
 & Facc.  & Uacc. & Facc. & Uacc.\\
\midrule
ECTC\cite{huang2016connectionist} & 27.7 & 35.6 & - & -\\
GRU reest.\cite{richard2017weakly} & 33.3 & - & - & - \\
TCFPN\cite{ding2018weakly} & 38.4 & - & 28.7 & - \\
NN-Viterbi\cite{richard2018neuralnetwork} & 43.0 & - & - & - \\
\hline
Ours w/o D${}^3$TW  & 34.9 & 36.1  &25.9 & 24.3 \\
Ours w/o Discriminative  & 38.0 & 38.4  & 30.0 & 28.3 \\
Ours (D${}^3$TW) & {\bf 45.7} & {\bf 47.4} & {\bf 33.6} & {\bf 30.5}\\
\bottomrule
\end{tabular}
\end{center}
\caption{
Weakly supervised action segmentation results in the Breakfast and Hollywood datasets. The use of both differentiable relaxation and discriminative modeling leads to the success of our \dttw and set our approach apart from previous approaches using ordering supervision.}
\vspace{-3mm}
\label{tab:segmentation}
\end{table}

\vspace{1mm}
{\noindent \bf Baselines.} We compare to the following six baselines:

{\noindent \it  - ECTC~\cite{huang2016connectionist}} does not rely on hard-EM. However, it uses non-differentiable DP based algorithm to compute its gradients. In addition, it does
include explicit models for the context between classes.

{\noindent \it  - GRU reest.~\cite{richard2017weakly}}  uses hidden Markov models and train their systems iteratively to reestimate the output. 

{\noindent \it  - TCFPN~\cite{ding2018weakly}}  is also based on action alignment. However, it uses an iterative framework that is neither differentiable nor discriminative like \dttw.

{\noindent \it  - NN-Viterbi~\cite{richard2018neuralnetwork}} is the most similar to ours, and can be seen as an ablation without discriminative modeling and without differentiable loss. However, our RNN takes the whole video as input instead of segments of the videos.

{\noindent \it  - Ours w/o \dttw} is our model without using \dttw but instead uses an iterative strategy similar to NN-Viterbi~\cite{richard2018neuralnetwork}. This ablation shows our model's performance without discriminative and differentiable modeling.

{\noindent \it  - Ours w/o Discriminative} is compared to show the importance of discriminative modeling for weakly supervised learning. Compared to \emph{Ours w/o \dttw}, this model use a differentiable relaxation of \eqnref{likelihood} as the objective. 

\begin{figure*}[tb]
\footnotesize
\centering
    \begin{tabular}{p{1.25cm}rccc}
    \textbf{Recipe} & \textbf{$\Delta$Facc.} & \textbf{Correct Predictions} & \textbf{False Positives} & \textbf{False Negatives} \\
     Sandwich & $+24.7\%$ &
    \raisebox{-.5\height}{\includegraphics[width=0.08\linewidth]{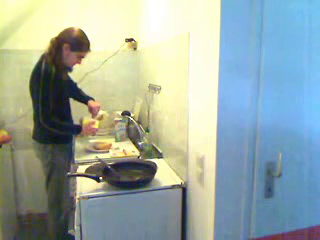}}
    \raisebox{-.5\height}{\includegraphics[width=0.08\linewidth]{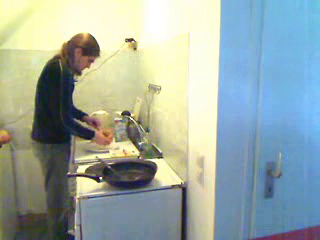}}
    \raisebox{-.5\height}{\includegraphics[width=0.08\linewidth]{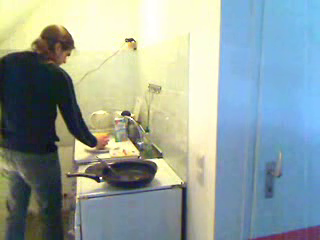}}&
    \raisebox{-.5\height}{\includegraphics[width=0.08\linewidth]{figures/qualitative/image4.png}}
    \raisebox{-.5\height}{\includegraphics[width=0.08\linewidth]{figures/qualitative/image5.png}}
    \raisebox{-.5\height}{\includegraphics[width=0.08\linewidth]{figures/qualitative/image6.png}} &
    \raisebox{-.5\height}{\includegraphics[width=0.08\linewidth]{figures/qualitative/image7.png}}
    \raisebox{-.5\height}{\includegraphics[width=0.08\linewidth]{figures/qualitative/image9.png}}
    \raisebox{-.5\height}{\includegraphics[width=0.08\linewidth]{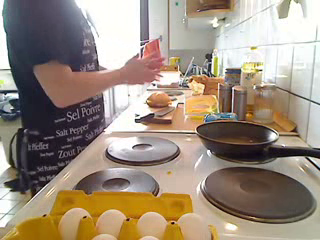}} \vspace{1pt} \\
    Cereals & $+19.9\%$ &
    \raisebox{-.5\height}{\includegraphics[width=0.08\linewidth]{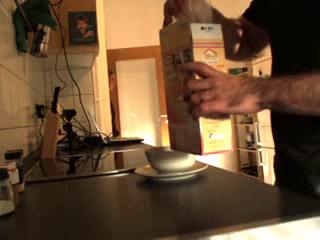}}
    \raisebox{-.5\height}{\includegraphics[width=0.08\linewidth]{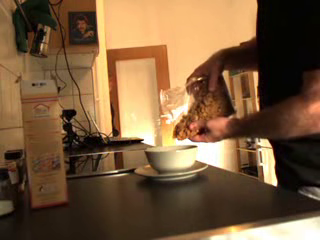}}
    \raisebox{-.5\height}{\includegraphics[width=0.08\linewidth]{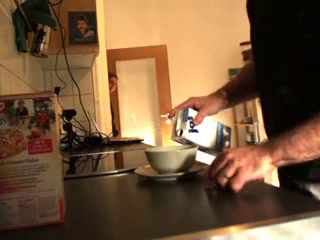}} &
    \raisebox{-.5\height}{\includegraphics[width=0.08\linewidth]{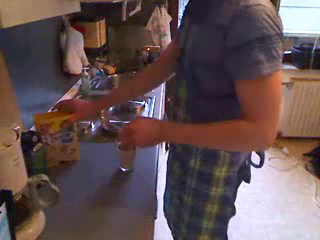}}
    \raisebox{-.5\height}{\includegraphics[width=0.08\linewidth]{figures/qualitative/image13.png}}
    \raisebox{-.5\height}{\includegraphics[width=0.08\linewidth]{figures/qualitative/image15.png}} &
    \raisebox{-.5\height}{\includegraphics[width=0.08\linewidth]{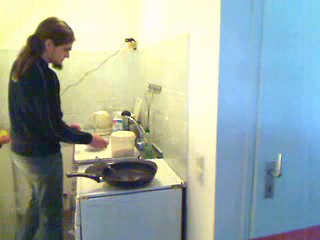}}
    \raisebox{-.5\height}{\includegraphics[width=0.08\linewidth]{figures/qualitative/image17.png}}
    \raisebox{-.5\height}{\includegraphics[width=0.08\linewidth]{figures/qualitative/image18.png}} \vspace{1pt} \\
    Pancake & $+0.2\%$ &
    \raisebox{-.5\height}{\includegraphics[width=0.08\linewidth]{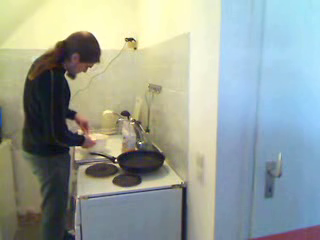}}
    \raisebox{-.5\height}{\includegraphics[width=0.08\linewidth]{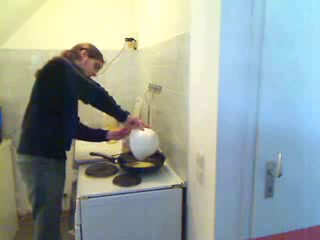}}
    \raisebox{-.5\height}{\includegraphics[width=0.08\linewidth]{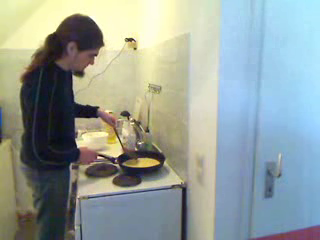}} &
    \raisebox{-.5\height}{\includegraphics[width=0.08\linewidth]{figures/qualitative/image22.png}}
    \raisebox{-.5\height}{\includegraphics[width=0.08\linewidth]{figures/qualitative/image23.png}}
    \raisebox{-.5\height}{\includegraphics[width=0.08\linewidth]{figures/qualitative/image24.png}}&
    \raisebox{-.5\height}{\includegraphics[width=0.08\linewidth]{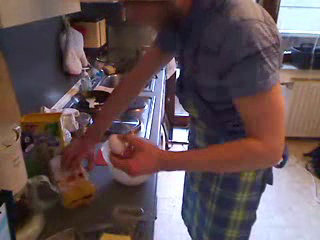}}
    \raisebox{-.5\height}{\includegraphics[width=0.08\linewidth]{figures/qualitative/image26.png}}
    \raisebox{-.5\height}{\includegraphics[width=0.08\linewidth]{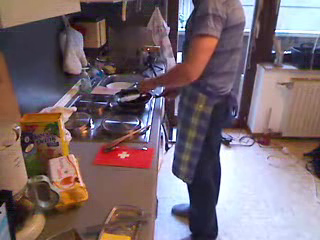}} \vspace{1pt} \\
    Scrambled Egg & $-0.8\%$ & 
    \raisebox{-.5\height}{\includegraphics[width=0.08\linewidth]{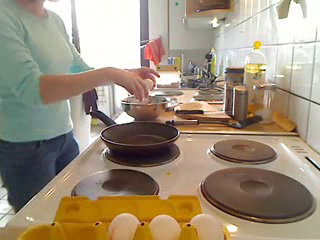}}
    \raisebox{-.5\height}{\includegraphics[width=0.08\linewidth]{figures/qualitative/image29.png}}
    \raisebox{-.5\height}{\includegraphics[width=0.08\linewidth]{figures/qualitative/image30.png}} &
    \raisebox{-.5\height}{\includegraphics[width=0.08\linewidth]{figures/qualitative/image31.png}}
    \raisebox{-.5\height}{\includegraphics[width=0.08\linewidth]{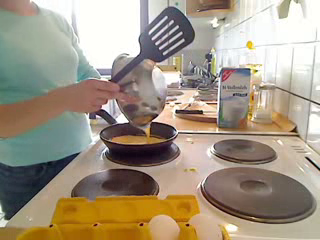}}
    \raisebox{-.5\height}{\includegraphics[width=0.08\linewidth]{figures/qualitative/image33.png}} & 
    \raisebox{-.5\height}{\includegraphics[width=0.08\linewidth]{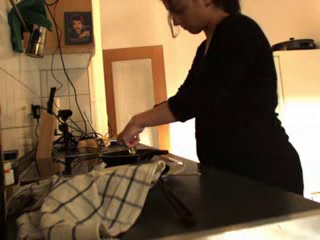}}
    \raisebox{-.5\height}{\includegraphics[width=0.08\linewidth]{figures/qualitative/image35.png}}
    \raisebox{-.5\height}{\includegraphics[width=0.08\linewidth]{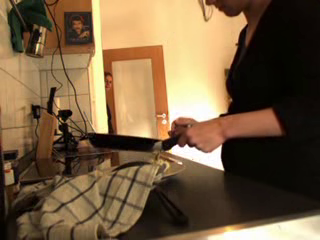}} \\
    \end{tabular}

   \vspace{+2mm}
   \caption{Qualitative results show the importance of discriminative modeling. We calculate $\Delta\text{Facc.}$, the absolute difference in frame accuracy between \emph{Ours \dttw} and \emph{Ours w/o Discriminative}. Discriminative modeling is able to improve the performances on almost all recipes or activities in the Breakfast dataset. In Pancake (row 3) and Scrambled Egg (row 4) where \dttw does not achieve a significant improvement, we see the challenge of cooking steps that are extremely similar from a further viewpoint. When cooking steps are distinct such as Sandwich (row 1) and Cereals (row 2), our \dttw is able to substantially improve the performance of frame accuracy by over 20\%.} 
   \vspace{-2mm}
\label{fig:dis_vis}
\end{figure*}

\subsection{Weakly Supervised Action Segmentation}

In the segmentation task, the goal is to predict frame-wise action labels for unseen test videos without any annotation. Weakly supervised action segmentation is challenging as the target output is never used in training.
As discussed in Section~\ref{sec:method_learning_inference}, we reduce the segmentation task to the alignment task by first finding the predicted transcript $\hat \ell$ that maximizes the likelihood in \eqnref{testtime} given a set of candidate transcripts $\lambda$, and then deriving the frame-wise labels from the alignment between $\hat \ell$ and video $X$. For a fair comparison, we follow~\cite{richard2018neuralnetwork} and set $\lambda$ to be the set of all transcripts seen in training time.

{\noindent \bf  Metrics.} 
We follow the metrics used in the previous work~\cite{kuehne2014language} to evaluate predicted frame-wise action labels. The first is \emph{frame accuracy}, the percentage of frames that are correctly labeled. The second is \emph{unit accuracy}, which is metric similar to the word error rate in speech recognition~\cite{klakow2002testing}. The output action label sequence is first aligned to the ground truth label sequence by vanilla dynamic time warping (DTW) before the error rate is computed. 

{\noindent \bf Results.} The results of weakly supervised action segmentation are shown in Table~\ref{tab:segmentation}. First, by explicitly modeling for the context between classes and their temporal progression, both GRU reest~\cite{richard2017weakly} and NN-Viterbi~\cite{richard2018neuralnetwork} are able to outperform ECTC by a large margin~\cite{huang2016connectionist}. In addition, we can see that using alignment is an effective strategy based on TCFPN~\cite{ding2018weakly}. \emph{Ours w/o \dttw} is able to combine these strengths and perform reasonably well compared to the state-of-the-art approaches. \emph{Ours w/o Discriminative} further improves on all metrics by using the differentiable relaxed loss function with better numerical stability. Most importantly, our full model using \dttw is able to combine the benefits of differentiable loss with discriminative modeling and significantly outperforms all the baselines and achieve state-of-the-art results on all metrics. This shows the importance of both components of our proposed \dttw model. Fig.~\ref{fig:vis} shows a qualitative comparison of models on a video making sandwich. Colors indicate different actions, and the horizontal axis is time. \emph{Ours \dttw} is the only model that correctly captures all the occurring actions with discriminative modeling. In addition, this also leads to more accurate boundaries of actions. Comparing \emph{NN-Viterbi} and \emph{Ours w/o Discriminative} shows the benefit of the differentiable model that leads to better action boundaries. 
In addition, we further illustrate the importance of discriminative modeling in Fig.~\ref{fig:dis_vis} by comparing our full model with \emph{Ours w/o Discriminative} and show the Correct Prediction, False Positives, and False Negatives of our model. As shown in the figure, discriminative modeling almost improves all 10 dishes in the Breakfast dataset, with  the only exception of Scrambled Egg that the \dttw is lower by a neglectable 0.2\% for the frame accuracy. We can see that for the dishes or activities of Pancake and Scrambled Egg that our \dttw does not improve much, the false positives are visually very similar to the correct prediction and lead to challenges of aligning the video with the transcript. On the other hand, for activities such as Sandwich and Cereals that involves distinct steps, our \dttw significantly improves the performance of the model by over 20\% of frame accuracy. In addition, if we look at the False Positives of Cereals, it is only fails because it is inherently difficult to distinguish visually similar actions of pouring cereals versus pouring flour from an obstructed viewing angle.

\subsection{Semi-Supervised Action Segmentation}

In contrast to most baselines, our formulation of weakly supervised action alignment based on DTW can easily incorporate any additional frame supervision by imposing path constraints in the calculation of $\psi_\gamma$. This is also called the frame-level semi-supervised setting, as proposed in~\cite{huang2016connectionist}. In semi-supervised setting, only a few frames in the video are sparsely annotated with the ground truth action, which is much easier for the annotator to annotate.

\begin{figure}[tb]
\centering
\includegraphics[width=1.0\linewidth]{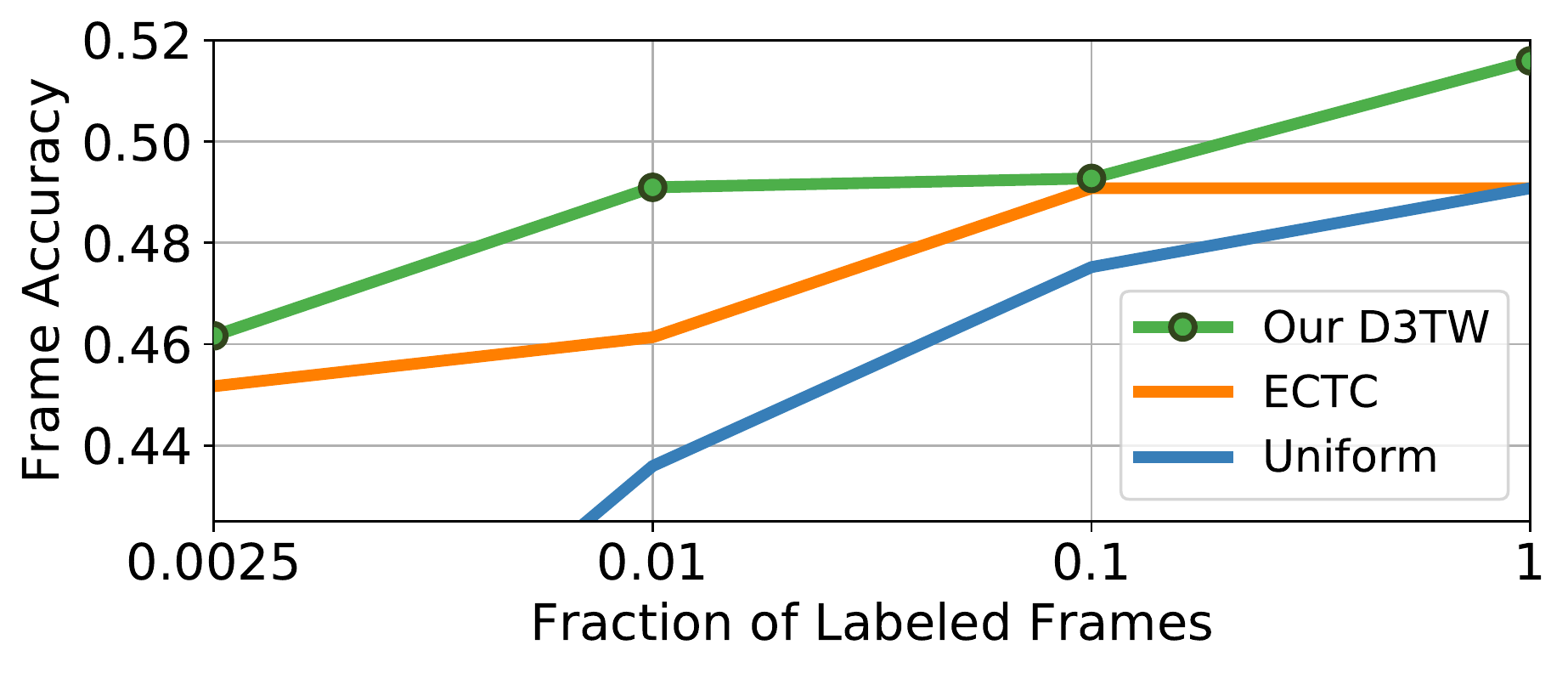}
\includegraphics[width=1.0\linewidth]{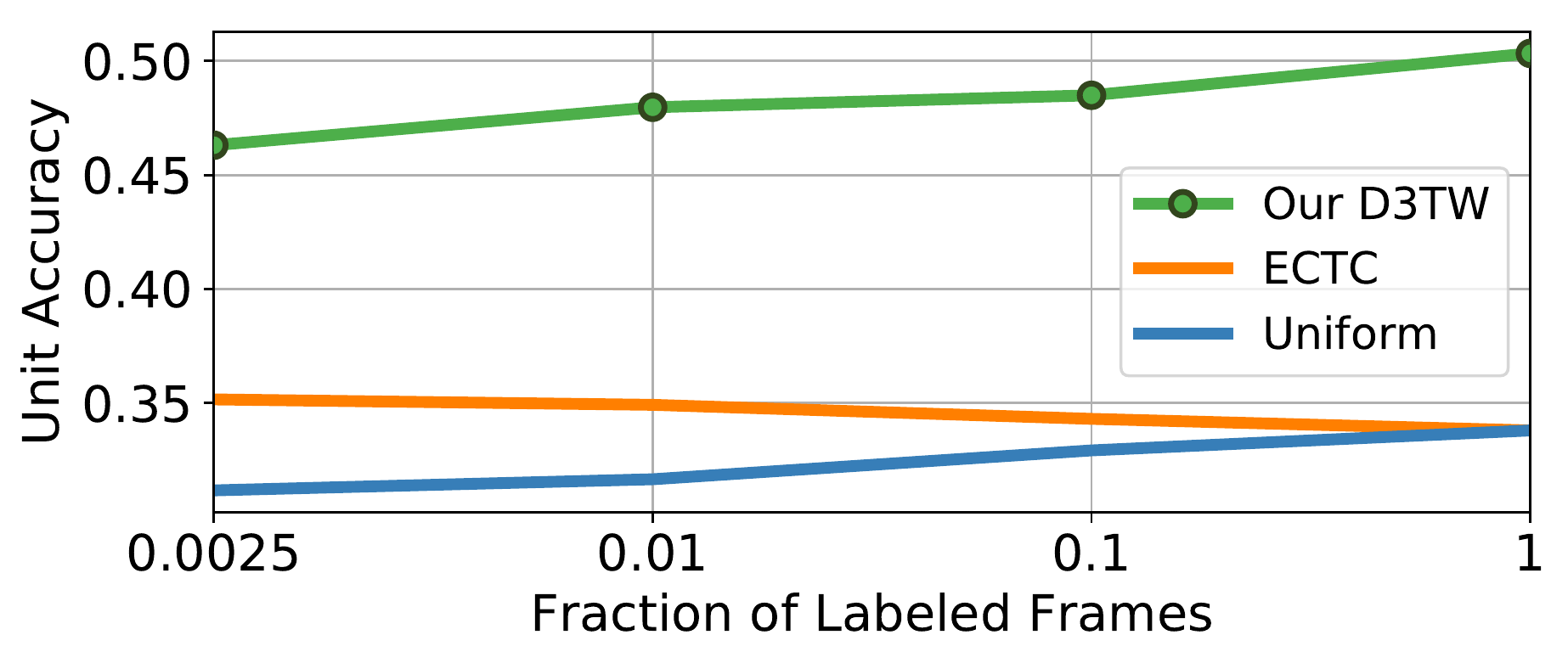}
\vspace{-2mm}
\caption{ Frame and unit accuracy are plotted against a fraction of labeled data in the frame-level semi-supervised setting for Breakfast dataset. Our DTW based formulation allows the frame-level supervision to be easily incorporated as the path constraints in dynamic programming. Our differentiable and discriminative modeling is able to lead to better performances on both metrics even in the semi-supervised setting.
}
\vspace{-3mm}
\label{fig:semi}
\end{figure}

In this setting, we only compare to ECTC as it is the only baseline that allows this experiment. We further compare to the ``Uniform'' baseline that was discussed in~\cite{huang2016connectionist}, where the model uses pseudo labels generated by uniformly distributing the transcript following the order. The results for frame-level semi-supervised action segmentation is shown in  Fig.~\ref{fig:semi}. We can see that the proposed \dttw is also able to  significantly improve performances in the semi-supervised setting. This again shows the importance of both the differentiable loss function and the discriminative modeling.

\subsection{Weakly Supervised Action Alignment}

In this task, the goal is to align the given transcript to its proper temporal location in the test video. Our \dttw formulation is designed to directly optimize for action alignment with only weak supervision. In this case, we always have the ground truth transcript $\ell^+$ and does not have to search using \eqnref{testtime}. It is noteworthy that the result from alignment can be interpreted as an empirical upper bound for our model's performance in action segmentation.

\begin{table}
\small
\begin{center}
\begin{tabular}{lcccc}
\toprule
& \multicolumn{2}{c}{Breakfast} & \multicolumn{2}{c}{Hollywood} \\
 & Facc.  & IoD & Facc. & IoD \\
\midrule
ECTC\cite{huang2016connectionist} (from \cite{ding2018weakly})  & $\sim$35  & $\sim$45 & - & $\sim$41 \\
GRU reest.\cite{richard2017weakly} & -  & 47.3 & - & 46.3 \\
TCFPN\cite{ding2018weakly} &53.5  & 52.3 & 57.4 &  39.6\\
NN-Viterbi\cite{richard2018neuralnetwork} & -  & - & - & 48.7\\
\hline
Ours w/o D${}^3$TW  & 42.8 & 49.5 & 51.2 & 47.2\\
Ours w/o Discriminative  & 52.3 & 47.6 & 51.8 & 46.9\\
Ours (D${}^3$TW) & {\bf 57.0} & {\bf 56.3} &{\bf 59.4} & {\bf 50.9}\\
\bottomrule
\end{tabular}
\end{center}
\caption{
Weakly supervised action alignment results. Compared to segmentation, the ground-truth transcript is given for the alignment, and thus the performances are higher. Nevertheless, both the differentiable relaxation and discriminative modeling are still beneficial for this task and lead to state-of-the-art results.
}
\vspace{-3mm}
\label{tab:aa}
\end{table}

{\noindent \bf  Metrics.} The primary goal of this experiment is to evaluate our model on aligning ground truth transcript to input video frames. We use metrics such as frame accuracy that measures the exact temporal boundaries in predictions. We drop unit accuracy as its use of DTW inevitably obfuscates the exact temporal boundaries. In addition to frame accuracy, we also measure the alignment quality with intersection over detection (IoD) following~\cite{bojanowski2014weakly}. Given a ground-truth action interval $I^*$ and a prediction interval $I$, IoD is defined as $\frac{|I \cap I^*|}{|I|}$. Readers should note that IoD is sometimes referred as Jaccard measure~\cite{bojanowski2014weakly,richard2018neuralnetwork}. The value of IoD is between 0 to 1 and the higher the better. We report the IoD averaged across all ground-truth intervals in the test set.

\vspace{1mm}
{\noindent \bf Results.} The results for weakly supervised action alignment are shown in Table~\ref{tab:aa}. We can see that the performance of all the baselines improves in terms of frame accuracy, this is because we have more information about the video in action alignment at test time. This also implies that the gap between different methods might be smaller. However, we observe the same trend as seen in action segmentation that the proposed \dttw is able to significantly outperform all the baselines on the metrics and achieve state-of-the-art result. This experiment once again validates that the use of both differentiable loss and discriminative modeling is important for our model's success.
\section{Conclusion}
\label{sec:conclusion}

We propose \dttw, the first discriminative framework for weakly supervised action alignment and segmentation. The key observation of our work is to use discriminative modeling between the positive and negative transcripts and bypass the problem of the degenerated sequence. The major challenge is that the dynamic programming based loss is often non-differentiable. We address this by proposing a continuous relaxation that allows \dttw to directly optimize for the discriminative objective with end-to-end training. Our results and ablation studies show that both the discriminative modeling and the differentiable relaxation are crucial for the success of \dttw, which achieves state-of-the-art results in both segmentation and alignment on two challenging real-world datasets. Our \dttw framework is general and can be extended to other tasks that require prior structures in the output and end-to-end differentiability.

\vspace{+2mm}
{\noindent \bf Acknowledgements.}
This work was partially funded by Toyota Research Institute (TRI). This article solely reflects the opinions and conclusions of its authors and not TRI or any other Toyota entity.

{\small
\bibliographystyle{ieee}

}

\end{document}